\documentclass[letterpaper]{article} 
\usepackage{aaai2026}

\usepackage{times}  
\usepackage{helvet}  
\usepackage{courier}  
\usepackage[hyphens]{url}  
\usepackage{graphicx} 
\usepackage{natbib}  
\usepackage{caption} 
\usepackage{algorithm}
\usepackage{algorithmic}

\usepackage{newfloat}
\usepackage{listings}
\DeclareCaptionStyle{ruled}{labelfont=normalfont,labelsep=colon,strut=off} 
\floatstyle{ruled}
\newfloat{listing}{tb}{lst}{}
\floatname{listing}{Listing}
\title{GBV-SQL: Guided Generation and SQL2Text Back-Translation Validation \\ for Multi-Agent Text2SQL}

\author{
  Daojun Chen\textsuperscript{\rm 1}\thanks{First Author}\quad
  Xi Wang\textsuperscript{\rm 2}\quad
  Shenyuan Ren\textsuperscript{\rm 3}\quad
  Qingzhi Ma\textsuperscript{\rm 1}\thanks{Corresponding Author}\quad
  Pengpeng Zhao\textsuperscript{\rm 1}\quad
  An Liu\textsuperscript{\rm 1}
}

\affiliations{
  \textsuperscript{\rm 1}School of Computer Science \& Technology, Soochow University, Suzhou, China\\
  \textsuperscript{\rm 2}Computer Science, University of Sheffield, UK\\
  \textsuperscript{\rm 3}School of Computer Science and Technology, Beijing Jiaotong University, China\\
  djchen@stu.suda.edu.cn, xi.wang@sheffield.ac.uk, syren@bjtu.edu.cn, qzma@suda.edu.cn, ppzhao@suda.edu.cn, anliu@suda.edu.cn
}

\usepackage{booktabs}
\usepackage{multirow}

\begin{document}

\maketitle

\begin{abstract}
While Large Language Models have significantly advanced Text2SQL generation, a critical semantic gap persists where syntactically valid queries often misinterpret user intent. To mitigate this challenge, we propose GBV-SQL, a novel multi-agent framework that introduces Guided Generation with SQL2Text Back-translation Validation. This mechanism uses a specialized agent to translate the generated SQL back into natural language, which verifies its logical alignment with the original question. Critically, our investigation reveals that current evaluation is undermined by a systemic issue: the poor quality of the benchmarks themselves. We introduce a formal typology for ``Gold Errors", which are pervasive flaws in the ground-truth data, and demonstrate how they obscure true model performance. On the challenging BIRD benchmark, GBV-SQL achieves 63.23\% execution accuracy, a 5.8\% absolute improvement. After removing flawed examples, GBV-SQL achieves 96.5\% (dev) and 97.6\% (test) execution accuracy on the Spider benchmark. Our work offers both a robust framework for semantic validation and a critical perspective on benchmark integrity, highlighting the need for more rigorous dataset curation.
\end{abstract}


\section{Introduction}
Text2SQL is a challenging task that involves automatically converting a Natural Language Question (NLQ) into Structured Query Language (SQL) that can be executed on a relational database \citep{Zelle1996Learning,qin2022survey}. The primary goal is to enable non-technical users to interact with complex databases using only natural language, removing the barrier of learning SQL syntax. As a long-standing objective in both natural language processing and database research \citep{nguyen2023vigptqa, zhu2023solving}, this field has been profoundly reshaped by recent advancements in Large Language Models (LLMs). These models, particularly through the paradigm of in-context learning (ICL), now represent the state-of-the-art, often achieving performance that surpasses traditional fine-tuned approaches \citep{pourreza2023din, li2023resdsql, liu2024survey}. 
\begin{figure}[t]
    \centering
    \includegraphics[width=0.96\columnwidth]{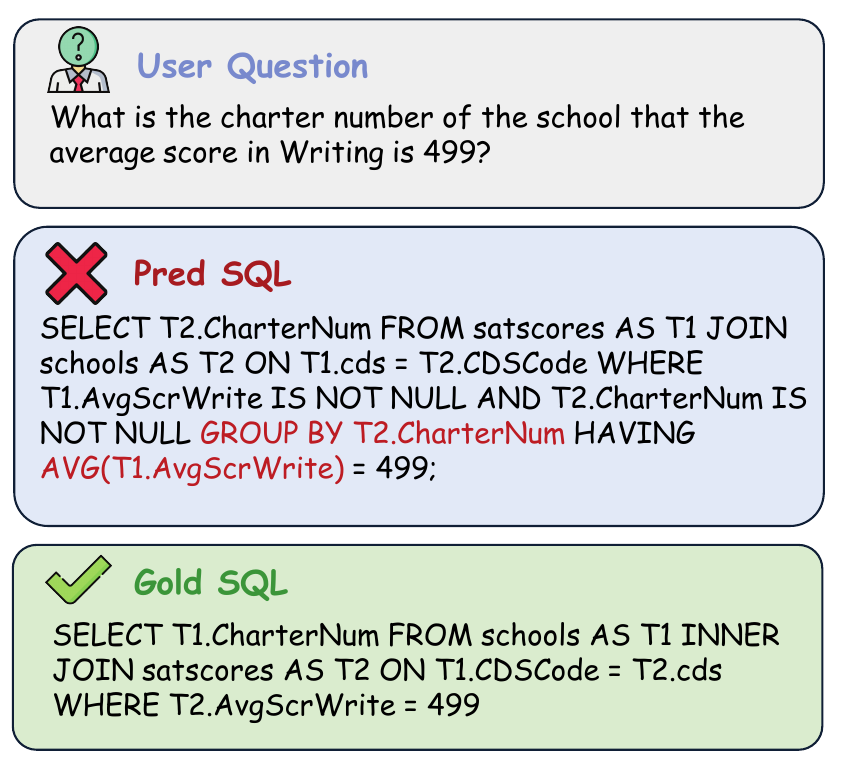}
    \caption{A BIRD dev example where syntactically valid Pred SQL semantically misinterprets the query: it unnecessarily aggregates ``average score" by charter groups, while the user explicitly seeks a single school with a score of 499.}
    \label{fig:SemErrorEx}
\end{figure}

\begin{figure*}[h]
    \centering
    \includegraphics[width=0.96\linewidth]{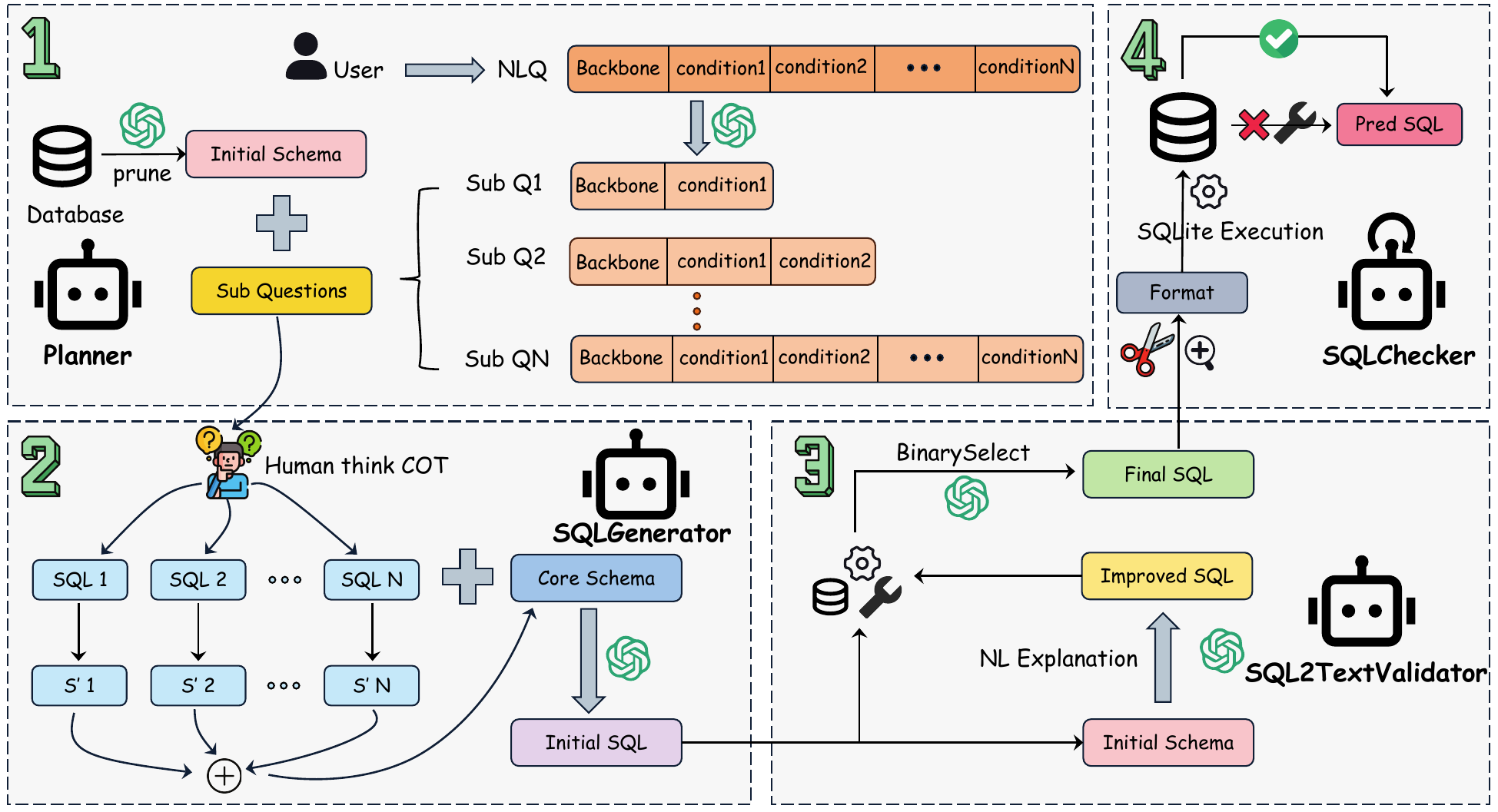}
    \caption{The workflow of GBV-SQL framework, including four agents: (1)\textit{Planner} (2)\textit{SQLGenerator} (3)\textit{SQL2TextValidator} (4)\textit{SQLChecker}.}
    \label{fig:GBV-SQL workflow}
\end{figure*}
Despite this significant progress, a critical challenge persists: ensuring the generated SQL query is not only syntactically correct but also semantically faithful to the user's original intent. This challenge stems from the inherent semantic gap between an unstructured Natural Language Question (NLQ) and the formal structure of SQL, which remains a primary source of failure. A syntactically valid query can still misinterpret the user's goal, producing plausible but erroneous results that are difficult to detect. For instance, as illustrated in Figure~\ref{fig:SemErrorEx}, when asked for a school whose ``average score'' column has a value of 499, a model might instead generate a query that performs an unnecessary aggregation, calculating the average of the ``average score'' column across a group of schools and checking if that new result is 499. This error, transforming a direct filtering condition (`WHERE') into a flawed aggregation condition (`HAVING'), highlights the inadequacy of syntax-only verification. While existing state-of-the-art methods employ techniques like query decomposition \citep{gao2024text, xie2024mag} and multi-agent collaboration \citep{wang2025mac} to ensure that LLMs generate reasonable SQL, they often lack a dedicated mechanism to robustly validate the final query's logic against the NLQ's semantics.

To bridge this semantic gap, we propose GBV-SQL, a multi-agent framework integrating Guided Generation with SQL2Text Back-translation Validation. The workflow is orchestrated by four specialized agents (Figure~\ref{fig:GBV-SQL workflow}). The process starts with a \textit{Planner} that prunes the database schema and decomposes the NLQ into simpler sub-problems. An \textit{SQLGenerator} then constructs the query using a novel Human-like Chain-of-Thought process. Critically, to ensure semantic fidelity, a dedicated \textit{SQL2TextValidator} performs back-translation, correcting the query to produce an improved version if a logical mismatch with the user's intent is found. Finally, an \textit{SQLChecker} conducts final checks, initiating an iterative repair cycle to resolve any syntactic or execution errors until the query is valid. This entire process is designed to systematically target and minimize inaccuracies.

Our rigorous evaluation on the Spider benchmark \citep{yu2018spider} yields a second, equally critical finding: a substantial portion of our model's apparent failures stems not from our method, but from inherent errors in the benchmark's ground-truth data. We term these issues ``Gold Errors'' and propose a formal typology for their systematic classification. This discovery fundamentally challenges the reliability of current evaluation practices and highlights a systemic issue in the field.

Our main contributions are threefold:
\begin{itemize}
    \item We propose GBV-SQL, a novel multi-agent framework that introduces a SQL2Text back-translation validation mechanism to explicitly bridge the semantic gap in Text2SQL, ensuring the generated query's logic is faithful to the user's original intent.
    \item We conduct a systematic analysis of errors in the widely-used Spider benchmark, introducing a novel typology for ``Gold Errors", which are flaws inherent in the ground-truth data itself.
    \item GBV-SQL achieves \textbf{63.23\%} execution accuracy on the challenging BIRD benchmark, a \textbf{5.8\%} absolute improvement. On a ``clean" version of Spider, corrected for identified ``Gold Errors", our model's execution accuracy reaches \textbf{96.5\%} (dev) and \textbf{97.6\%} (test), revealing its true capabilities and the critical impact of benchmark quality.
\end{itemize}

\section{Related Work}
\subsubsection{ICL-based Methods for Text2SQL}
In-context learning (ICL) with Large Language Models (LLMs) is the current state-of-the-art paradigm for Text2SQL. A dominant strategy for tackling complex queries is decomposition, where methods break a difficult question into simpler sub-problems. For example, DIN-SQL \cite{pourreza2023din} first applied CoT prompting for decomposition.  
DAIL-SQL \cite{gao2024text} then explored advanced prompt designs in this framework.  
Tai et al. evaluated multiple prompt strategies and introduced QDecomp to incrementally add schema details to sub-questions \cite{tai2023exploring}.
To improve overall robustness, other research has focused on two primary avenues. The first is enhancing the quality of the model's input via more accurate schema linking \citep{cao2024rsl, wang2025linkalign, li2024pet}. The second is refining the model's output through post-hoc correction of generated SQL \citep{dong2023c3, ren2024purple} or by selecting the optimal query from multiple candidates \citep{lee2025mcs, pourreza2024chase}. While these diverse strategies have advanced the field, they predominantly operate in a unidirectional workflow. They lack an explicit mechanism to verify if the final, syntactically correct SQL is truly semantically aligned with the user's original intent. Our work directly targets this semantic gap with a novel validation step.

\subsubsection{Multi-Agent Systems for Text2SQL}
To orchestrate the multi-step reasoning process, recent work has turned to multi-agent systems. MAC-SQL \citep{wang2025mac} and MAG-SQL \citep{xie2024mag} pioneered this approach by employing several LLM-based agents to collaboratively manage decomposition, SQL generation, and refinement. These frameworks demonstrated that a division of labor can effectively handle complex queries. However, existing agents only collaborate to generate SQL; none perform an explicit semantic verification of the final query. Other agent-based systems, like SQLFixAgent \citep{cen2025sqlfixagent}, focus specifically on correcting errors from fine-tuned models, which is a different task context. Our GBV-SQL framework introduces a new agent archetype: the SQL2TextValidator. This agent acts as a skeptical verifier whose sole purpose is to perform SQL-to-Text back-translation. By challenging the semantic fidelity of the generated query from an inverse perspective, it provides a critical feedback loop absent in prior collaborative models.

\subsubsection{Benchmark Quality and Gold Errors}
The reliability of Text2SQL evaluation is intrinsically tied to the quality of benchmark datasets. The presence of flaws in the ground-truth labels, termed ``Gold Errors", has been noted in the error analyses of prior work \citep{wang2025mac, lee2025mcs}. Further research has delved into specific quality issues, such as analyzing label accuracy \citep{renggli2025fundamental}, classifying question ambiguity \citep{dong2024practiq}, or automatically detecting incorrect SQL-NLQ mappings \citep{yang2025automated}. These efforts highlight a growing awareness of data quality challenges. Our work builds upon this foundation by proposing a formal, structured typology for Gold Errors. Unlike previous post-hoc analyses, our classification system is more comprehensive, categorizing errors originating from the SQL, the NLQ, and the database itself. This systematic framework not only allows for a more precise and fair re-evaluation of GBV-SQL's true capabilities but also offers a valuable tool for the future curation and maintenance of Text2SQL benchmarks.

\begin{figure}[t]
    \centering
    \includegraphics[width=0.96\linewidth]{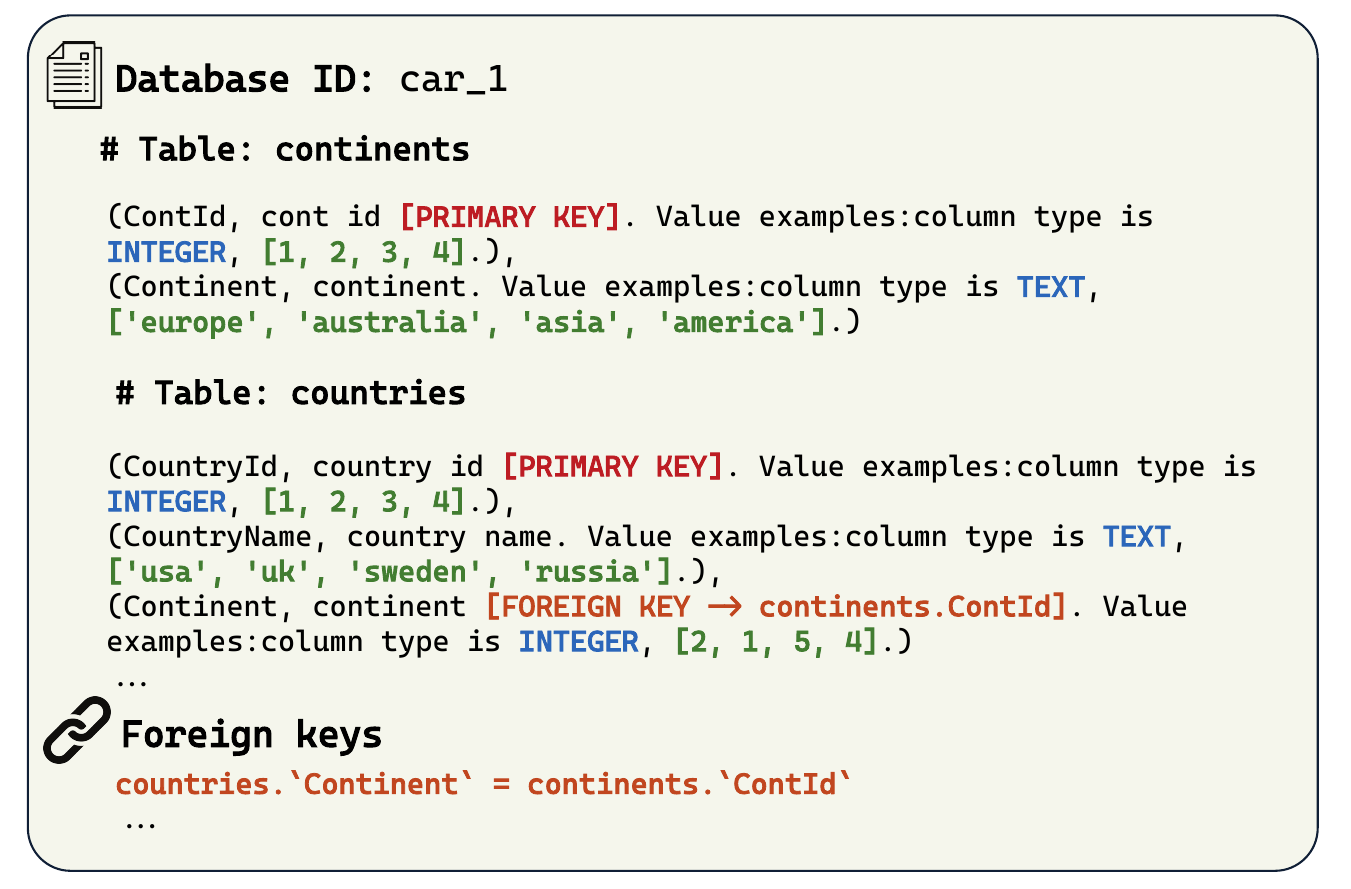}
    \caption{The prompt template for Schema Representation.}
    \label{fig:Schema}
\end{figure}
\section{Methodology}
\subsection{Overview}
In this section, we introduce GBV-SQL, a novel multi-agent framework for Text2SQL leveraging large language models. As shown in Figure~\ref{fig:GBV-SQL workflow}, our framework comprises four agents: \textit{Planner}, \textit{SQLGenerator}, \textit{SQL2TextValidator} and \textit{SQLChecker}. The Planner initiates the process by invoking an LLM to prune the database schema to its most relevant elements and then decomposes the NLQ into sub-questions. 
Mimicking human thought processes, the SQLGenerator generates SQL for each sub-question before synthesizing them into the final query. The SQL2TextValidator ensures semantic integrity by prompting an LLM to translate the complete SQL back into natural language for comparison with the original NLQ, yielding a refined query. Finally, the SQLChecker conducts format and syntax checks, correcting any errors to guarantee an executable final SQL.

\subsection{Planner}
\subsubsection{Schema Representation}
Prior work \citep{bogin2019representing, guo2019towards, lei2020re, gao2024xiyan} has shown that a well-organized representation of the database schema can improve the accuracy of SQL generation by LLMs. In this paper, we organize our schema description based on the approach in MAC-SQL, but with augmentations. We incorporate the data types of database columns and provide more intuitive foreign key descriptions. Furthermore, we are the first to include descriptions for primary keys within the schema representation. The specific prompt design is illustrated in Figure~\ref{fig:Schema}.

\subsubsection{Schema Pruning}
Overly long schemas can cause an LLM to focus on information irrelevant to the NLQ and incur high token costs. Therefore, we perform an initial pruning of the full database schema by prompting the LLM. We prompt the LLM to analyze the NLQ ($Q$) and the full database schema ($D$), retaining only the most relevant information.
\begin{equation}
  \label{eq:S_initial}
  S_{\mathrm{initial}} = \mathrm{Prune}(Q, D)\,,
\end{equation}
where $S_{\mathrm{initial}}$ represents the pruned schema relevant to the question.

\subsubsection{NLQ Decomposition}
As demonstrated in prior surveys \cite{pourreza2023din, pourreza2024dts}, decomposing a question into sub-questions is an effective strategy. We employ a decomposition strategy based on the divide-and-conquer paradigm, wherein question decomposition and SQL generation are separate processes. Specifically, we adopt the Targets-Conditions method proposed in MAG-SQL \cite{xie2024mag} to decompose the NLQ based on its distinct conditional clauses and query targets, as shown in Figure~\ref{fig:GBV-SQL workflow}.
\begin{figure}[t]
    \centering
    \includegraphics[width=0.96\columnwidth]{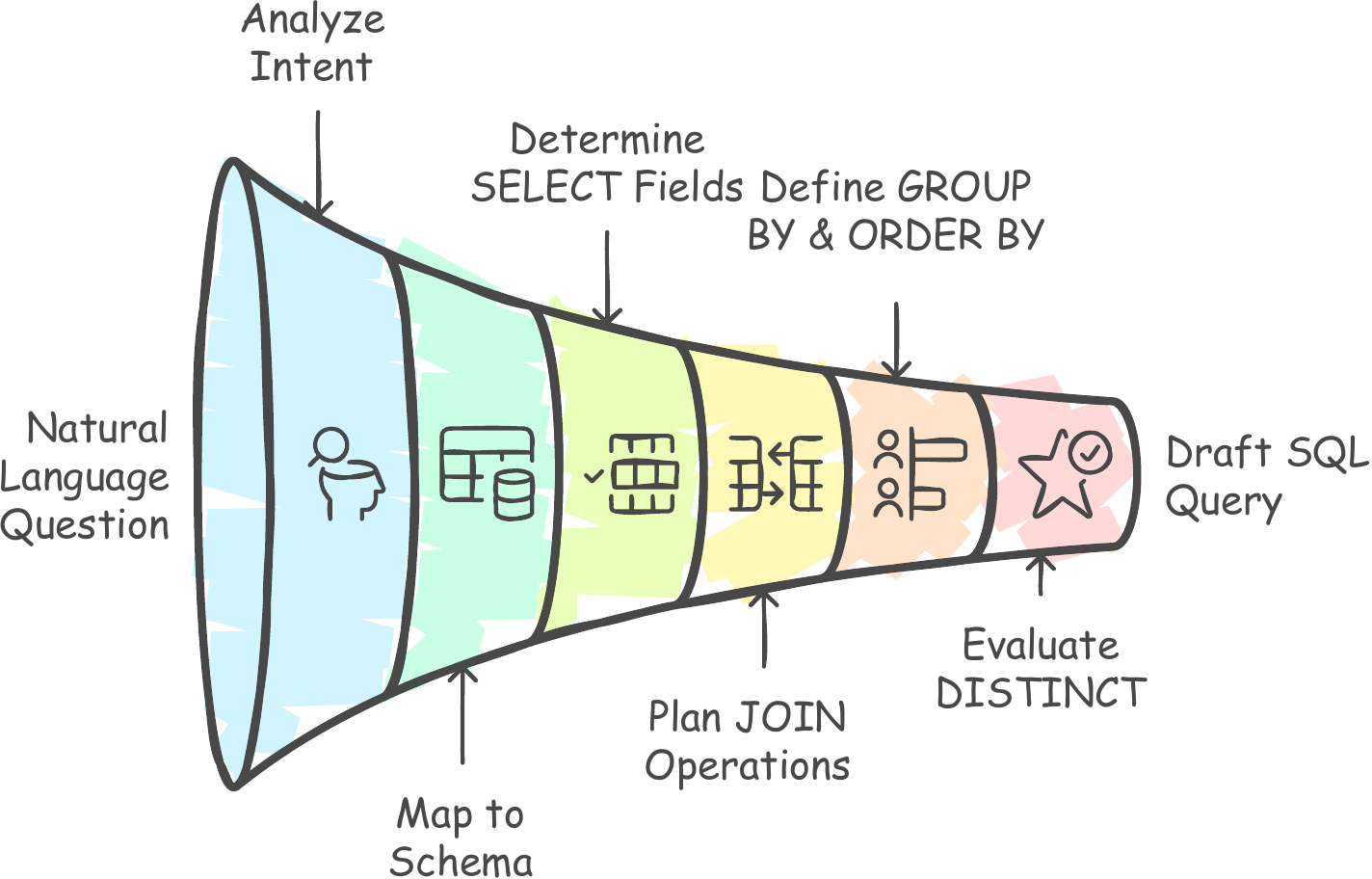}
    \caption{Human-like Chain-of-Thought Workflow for Drafting SQL Queries.}
    \label{fig:HumanCoT}
\end{figure}
\subsection{SQLGenerator}
\subsubsection{Human-like CoT}
To emulate the human thought process for drafting SQL queries \cite{yuan2025cogsql}, we have designed a novel CoT method, illustrated in Figure~\ref{fig:HumanCoT}. This approach first analyzes the intent of the NLQ. It then identifies the relevant tables and columns based on the database schema and determines the specific information to be returned. Subsequently, it systematically analyzes how to construct essential SQL clauses, considering the specific logic for joins, the columns for grouping and ordering, and the potential need for deduplication with DISTINCT. Finally, it emphasizes adherence to proper SQL syntax throughout the generation process.

\subsubsection{Sub-SQL Generation and Merging}
Our framework employs a two-stage process to generate the SQL. First, for each sub-question decomposed by the Planner, we generate a corresponding sub-SQL. Subsequently, we merge the contextual information from these sub-SQLs to synthesize the final answer. This process begins by applying our Human-like CoT method to generate a sub-SQL for each sub-question ($q_i$) using the initially pruned schema, $S_{\mathrm{initial}}$. A crucial step follows where we perform a ``back-linking'' operation on each generated sub-SQL ($SQL_i$). This operation is an LLM inference step, where the model is prompted to extract the precise set of table-column pairs ($S_i$) that were essential for the sub-query's creation.
\begin{equation}
  \label{eq:sql_humancot}
  SQL_i = \mathrm{HumanCoT}\bigl(S_{\mathrm{initial}},\,q_i\bigr)
\end{equation}
\begin{equation}
  \label{eq:si_backlink}
  S_i = \mathrm{BackLink}\bigl(SQL_i\bigr)
\end{equation}
These extracted schema subsets are then merged to create a more refined and highly relevant schema, $S_{\mathrm{core}}$. This new schema is more accurate than $S_{\mathrm{initial}}$ because it is derived directly from the SQL logic required to answer each sub-question. The merging operation is defined as the union of all table-column pairs from the individual schema subsets:
\begin{equation}
    S_{\mathrm{core}} = \bigoplus_{i=1}^n S_i
\end{equation}
where the operator is defined for any table $t_k$ in the schema as:
\begin{equation}
    \left(\bigoplus_{i=1}^n S_i\right)(t_k) = \bigcup_{i=1}^n S_i(t_k)
\end{equation}
Finally, this consolidated schema $S_{\mathrm{core}}$, along with the complete set of sub-question and sub-SQL pairs, provides a rich and focused context for the LLM to generate the final comprehensive query, referred to as Initial SQL in Figure~\ref{fig:GBV-SQL workflow}.

\subsection{SQL2TextValidator}
As one of the core modules in GBV-SQL, the SQL2TextValidator performs semantic validation by prompting the LLM to translate the initial SQL query from the SQLGenerator into a detailed natural language explanation. This explanation describes the query's overall function and the specific mechanics of its components, with a particular focus on the query's execution process. The LLM is then tasked with comparing this explanation against the original NLQ to identify any semantic discrepancies. If the logic is consistent, the query is accepted; otherwise, the LLM is prompted to correct the SQL to align with the user's intent, yielding a validated improved SQL. To maximize the validation's efficacy, we incorporate a binary selector inspired by RSL-SQL \cite{cao2024rsl}. This selector uses the LLM to evaluate the semantic consistency of both the initial and the validated SQL versions against the original NLQ, taking their execution results into account to select the Final SQL.

\subsection{SQLChecker}
Due to the hallucinatory nature of large models, the generated SQL can often contain formatting errors \cite{shen2025study}. Therefore, this module introduces an SQL format trimmer. This component optimizes the query according to two key principles: a minimization principle, which eliminates superfluous fields and redundant operations, and a minimal usability principle, which avoids intrusive formatting (e.g., rounding decimals) that could alter the result's meaning unless explicitly required. Subsequently, we employ a multi-round iterative checking process to analyze the query's executability. If the SQL execution yields an empty result, a value of 0, or contains NULL values, the query is repaired. This repair cycle continues until the query executes successfully or the maximum number of iterations is exceeded. For the repair process, we retrieve the top-k most relevant values from the database based on the SQL and combine them with the partial execution results as context for prompting the LLM to make corrections. Algorithm~\ref{alg:SQLChecker} details this process.
\begin{algorithm}[tb]
  \caption{The SQLChecker Algorithm}
  \label{alg:SQLChecker}
  \textbf{Input:} Question $q$, database $db$, schema $S'$, initial SQL \\
  \hspace*{1.5em}$sql_{\mathrm{pre}}$\\
  \textbf{Output:} final SQL $sql$
  \begin{algorithmic}[1]
    \STATE $sql \gets \texttt{ReduceFormat}(q,\,sql_{\mathrm{pre}},\,S')$
    \STATE $count \gets 0$
    \WHILE{$count < \mathrm{maxTryTime}$}
      \STATE $(\mathit{pass},\,err,\,res)\gets \texttt{execTool}(sql,\,db)$
      \IF{$\mathit{pass}$}
        \STATE \textbf{break}
      \ELSE
        \STATE $vals\gets \texttt{valueRetrieve}(q,\,sql,\,S',\,db)$
        \STATE $sql\gets \texttt{Refiner}(q,\,sql,\,S',\,err,\,res,\,vals)$
      \ENDIF
      \STATE $count\gets count + 1$
    \ENDWHILE
    \STATE \RETURN $sql$
  \end{algorithmic}
\end{algorithm}

\section{Experiments}
\subsection{Experimental Setup}

\begin{table*}[t]
  \centering
  {\small  
  \begin{tabular}{c|cccc|c}
    \toprule
    \multirow{2}{*}{\textbf{Methods}} & \multicolumn{4}{c|}{\textbf{EX}} & \multirow{2}{*}{\textbf{VES}} \\
    & \textbf{Simple} & \textbf{Moderate} & \textbf{Challenging} & \textbf{Total} & \\
    \midrule
    GPT-4 (zero-shot)       & --      & --      & --      & 46.35 & 51.75 \\
    Deepseek-v3 (zero-shot) & 49.84   & 34.70   & 25.52   & 42.96 & 53.25 \\
    DIN-SQL + GPT-4         & --      & --      & --      & 50.72 & 58.79 \\
    MAG-SQL + GPT-4         & --      & --      & --      & 61.08 & -- \\
    DAIL-SQL + GPT-4        & 63.02   & 45.59   & 43.05   & 54.76 & -- \\
    CogSQL + GPT-4    & --   & --   & --   & 59.58 & 64.30 \\
    CogSQL + Deepseek-v2    & 64.11   & 46.67   & 39.58   & 56.52 & -- \\
    MAC-SQL + GPT-4         & 65.73   & 52.96   & 40.28   & 59.39 & 66.24 \\
    \midrule
    MAC-SQL + Deepseek-v3   & 62.92   & 50.75   & 43.75   & 57.43 & 68.40 \\
    GBV-SQL + Deepseek-v3 (ours)
      & \textbf{69.51}
      & \textbf{54.62}
      & \textbf{50.69}
      & \textbf{63.23} ($\uparrow$5.8\%)
      & \textbf{69.87} \\
    \bottomrule
  \end{tabular}
  }
  \caption{\label{tab:bird-ex-ves}
    The EX and VES results on BIRD's dev set.
  }
\end{table*}

\subsubsection{Datasets}
We evaluate GBV-SQL on two widely-used Text2SQL benchmarks: Spider \cite{yu2018spider}, to assess cross-domain generalization across its over 200 databases (e.g., \texttt{flight\_2}), and the more challenging BIRD \cite{li2023can}, to test performance against real-world complexity involving noisy data and external knowledge. While Spider 2.0 \citep{lei24112} is an important industry-oriented evolution, its complexity in long-context understanding and multi-tool usage currently hinders a clear evaluation of our semantic validation mechanism. We plan to adapt GBV-SQL for such enterprise scenarios in future work.
\subsubsection{Baseline}
In our experiments, we compare our proposed method GBV-SQL against a wide range of ICL-based methods, including DIN-SQL \cite{pourreza2023din}, DAIL-SQL \cite{gao2024text}, MAC-SQL \cite{wang2025mac}, MAG-SQL \cite{xie2024mag}, SuperSQL \cite{li2024dawn}, and CogSQL \cite{yuan2025cogsql}.
\subsubsection{Evaluation Metrics}
We evaluate our framework using two metrics: execution accuracy and valid efficiency score. \textbf{\textit{Execution accuracy (EX)}} is defined as the proportion of predicted SQL queries whose execution results exactly match those of the ground-truth SQL. The \textbf{\textit{valid efficiency score (VES)}}, introduced by the BIRD dataset, jointly considers the correctness and execution efficiency of generated SQL. The efficiency score (related to execution time) is calculated only if the predicted and real SQL results match; otherwise, the score is 0.

\subsubsection{Implementation Details}
Our experiments primarily use Deepseek-v3 as the backbone large language model, accessed through the Deepseek API. For supplementary validation, we also employ GPT-4o via the OpenAI API. The temperature for both models is set to 0. In our SQLChecker module, the maximum number of repair iterations is set to 3. Our multi-agent architecture draws inspiration from MAC-SQL. Our method is capable of handling moderately complex queries, covering scenarios that involve both single and multiple tables.
\begin{table}[t]
  \centering
  {\small  
  \begin{tabular}{lcc}
    \toprule
    \textbf{Methods} & \textbf{Dev} & \textbf{Test} \\
    \midrule
    GPT-4 (zero-shot)            & 74.6 & --    \\
    DIN-SQL + GPT-4              & 82.8 & 85.3 \\
    DAIL-SQL + GPT-4             & 84.4 & 86.6 \\
    CogSQL + GPT4                & 85.4 & 86.4 \\
    SuperSQL + GPT4                & 87.0 & -- \\
    MAG-SQL + GPT-4        & 85.3 & 85.6 \\
    MAC-SQL + GPT-4              & 86.8 & 82.8 \\
    \midrule
    MAC-SQL + Deepseek-v3        & 80.1 & 77.6 \\
    \textbf{GBV-SQL + Deepseek-v3 (ours)} & 79.6 & 82.8($\uparrow$5.2\%) \\
    \textbf{GBV-SQL + GPT-4o (ours)} & 79.7 & 83.9 \\
    \textbf{GBV-SQL + No Gold Errors} & \textbf{96.5} & \textbf{97.6} \\
    \bottomrule
  \end{tabular}
  }
  \caption{The EX results on the Spider dev and test sets. Improvements relative to MAC-SQL+Deepseek-v3 are shown in parentheses. The ``No Gold Errors" designation indicates evaluation on a cleaned subset of the benchmark, from which entries with quality issues have been removed.}
  \label{tab:spider-result}
\end{table}

\subsection{Main Result}
\subsubsection{BIRD Result}
The results in Table~\ref{tab:bird-ex-ves} show that our method significantly enhances the performance of the Deepseek baseline. Using the Deepseek-v3 model, GBV-SQL achieves an EX of 63.23\% on the BIRD dev dataset, which is a 5.8\% improvement over MAC-SQL that also uses the Deepseek model, and even surpasses numerous methods that use GPT-4. This highlights the effectiveness of GBV-SQL in Text2SQL tasks.

\subsubsection{Spider Result}
Table~\ref{tab:spider-result} shows the performance of our framework and baseline methods on the Spider dataset. GBV-SQL achieves execution accuracies of 79.6\% and 82.8\% on the Spider dev and test sets, respectively. On the test set, our method outperforms the MAC-SQL baseline by a notable 5.2\% when using the same Deepseek-v3 model, achieving performance comparable to methods that rely on GPT-4. However, the performance on the dev set is unexpectedly lower than anticipated. This discrepancy motivates a deeper investigation into the nature of the execution failures, which reveals that the issue stems not from our model's capabilities but from what we identify as ``Gold Errors": pervasive quality issues within the benchmark's ground-truth data itself, as detailed in the following section.
\begin{figure*}[ht]
    \centering
    \includegraphics[width=0.96\linewidth]{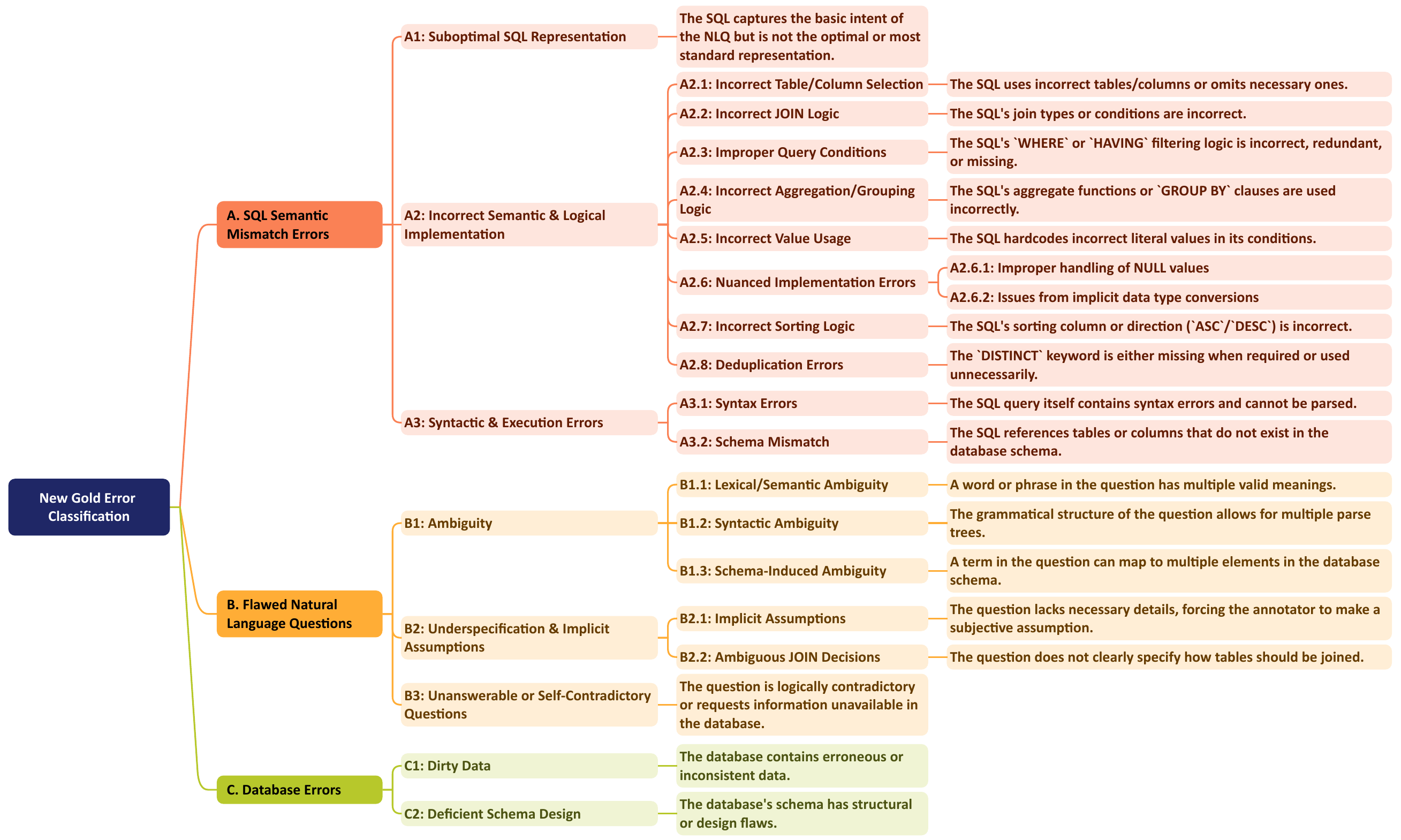}
    \caption{The new Gold Error Classification.}
    \label{fig:GoldError}
\end{figure*}
\subsection{Error Analysis}
A rigorous manual review of all items that failed execution on the Spider benchmark reveals the overwhelming majority of failures are not caused by our model but by inherent quality issues in the dataset. Consequently, we develop a novel framework for classifying these gold-standard errors.
\subsubsection{Gold Error Identification}
To ensure an objective and reproducible analysis, we establish a rigorous protocol to audit the quality of each data item. The process is conducted by three SQL-proficient graduate students. Initially, two students independently inspect and classify potential quality issues in every data item according to our proposed typology (Figure~\ref{fig:GoldError}), achieving a substantial inter-annotator agreement (Cohen's Kappa = 0.86). A third student then adjudicates all disagreements to make the final classification, ensuring the reliability of our findings.

\subsubsection{A New Typology of Gold Errors}
While prior research has acknowledged the existence of data quality problems, a clear and systematic classification has been absent. We address this gap by introducing a new standard that categorizes Gold Errors into three distinct types based on a comprehensive analysis of recurring issues:

\begin{itemize}
  \item \textbf{Type A (SQL-Side Errors)}: Covers all errors originating from the gold SQL itself, assuming the NLQ is valid. This includes semantic errors where the query fails to correctly or optimally represent the user's intent, as well as syntactic errors that prevent the query from executing.

  \item \textbf{Type B (NLQ-Side Errors)}: Includes all cases where the NLQ itself is the source of the problem. This covers questions that are ambiguous, underspecified, logically flawed, or unanswerable with the given database.

  \item \textbf{Type C (Database Errors)}: Signifies errors found within the database itself, such as flawed schema design or inconsistent and ``dirty'' data values.
\end{itemize}
As detailed in Figure~\ref{fig:GoldError}, each primary category is further partitioned into a granular hierarchy of subcategories.

\subsubsection{Analysis of Gold Errors in Spider}
\begin{figure}[t]
    \centering
    \includegraphics[width=0.98\columnwidth]{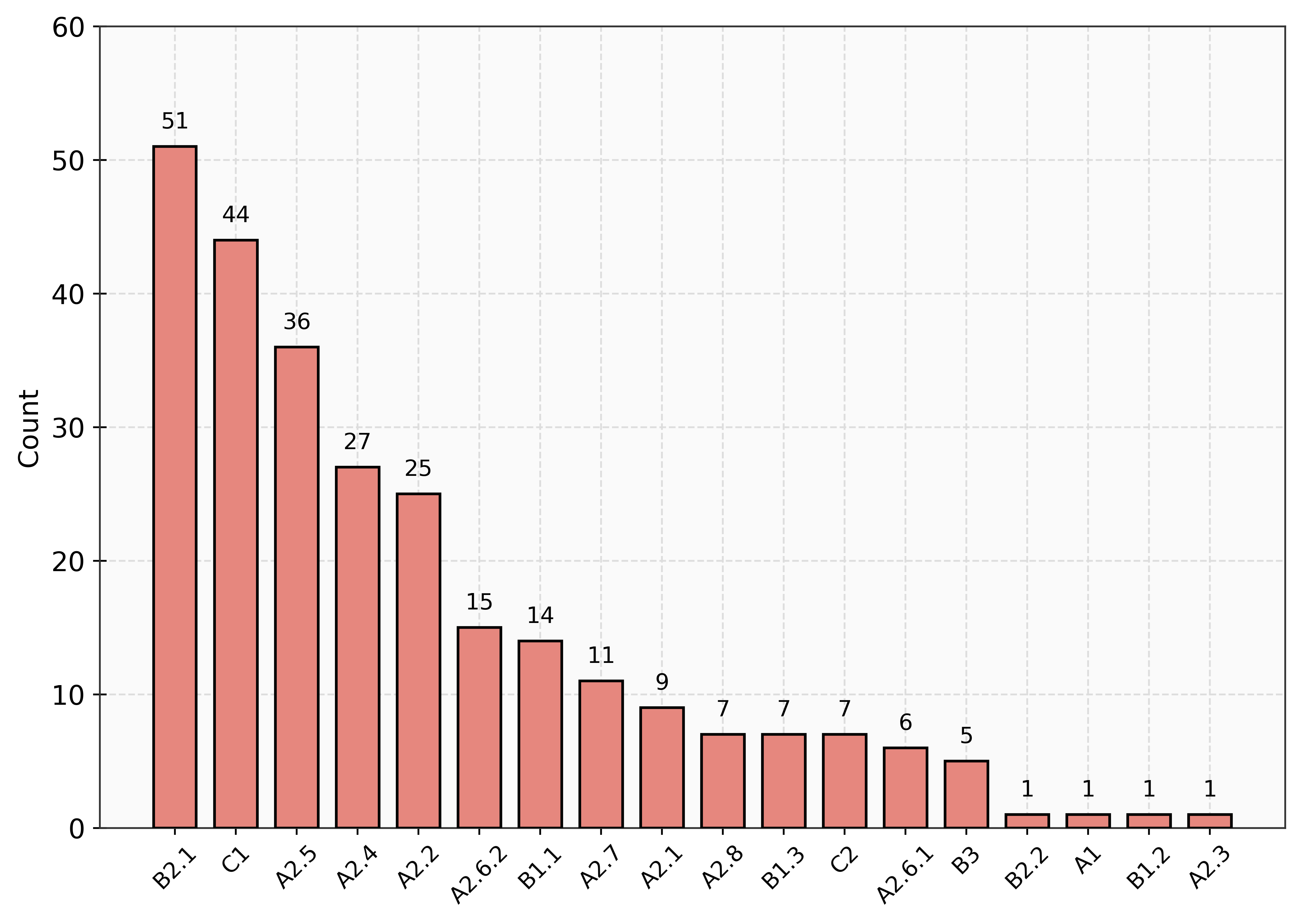}
    \caption{The distribution of Gold Error types in the Spider's dev set with EX=0. }
    \label{fig:SpiderError}
\end{figure}

\begin{figure}[t]
    \centering
    \includegraphics[width=0.98\columnwidth]{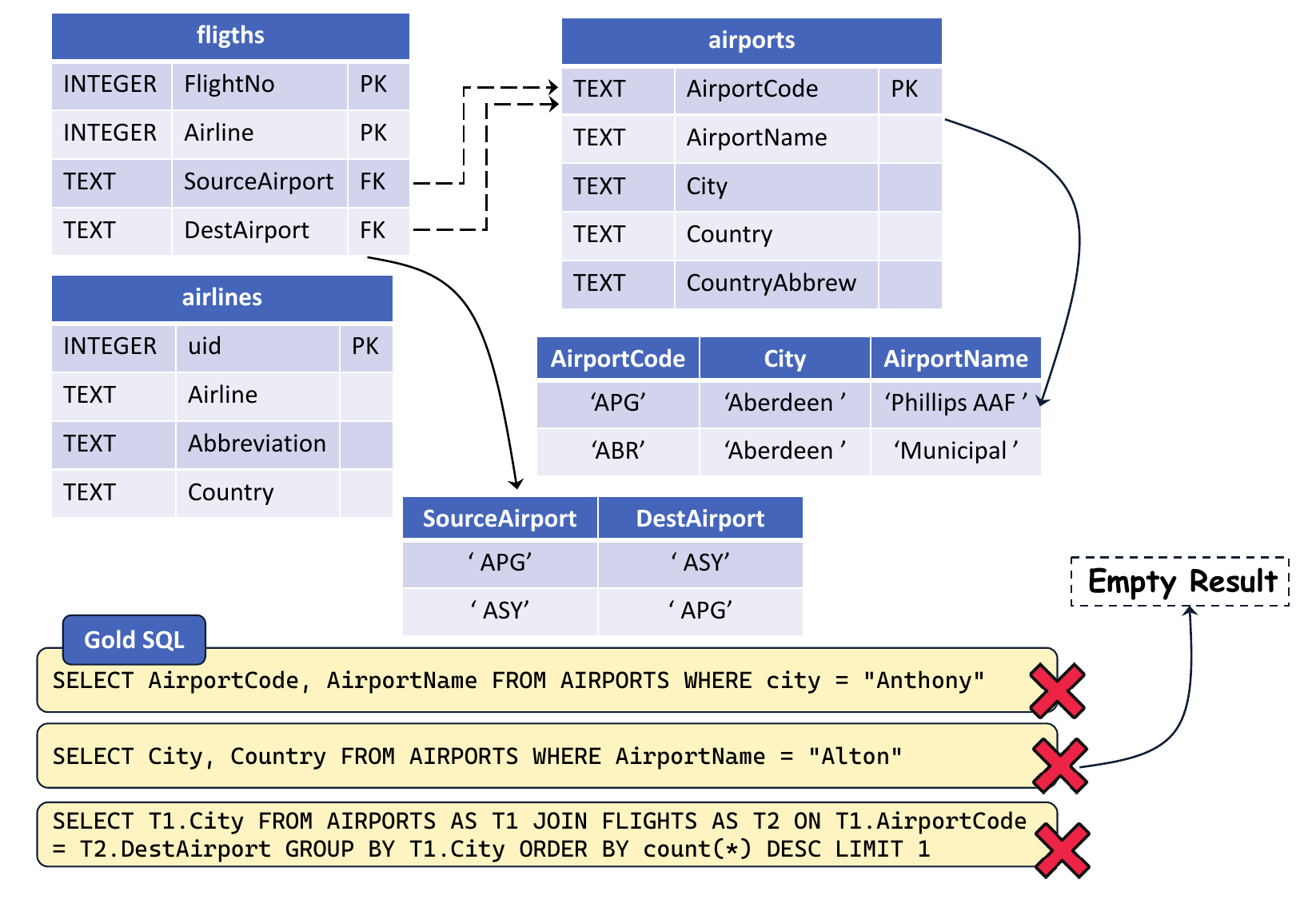}
    \caption{An example of (C1:Dirty Data) from database \texttt{flight\_2} in Spider's dev set.}
    \label{fig:DatabaseError}
\end{figure}

Our analysis of the Spider dev set uncovers a significant number of ``Gold Errors" (i.e., errors in the ground-truth labels), with 183 instances found in samples our model failed (EX=0) and 62 in those it passed (EX=1). Figure~\ref{fig:SpiderError} provides a quantitative breakdown of the various Gold Error categories within these failed samples. A prominent example is the data contamination within the \texttt{flight\_2} database, which is illustrated in Figure~\ref{fig:DatabaseError}; here, extraneous spaces in TEXT attributes cause many official Gold SQL queries to fail, whereas our method correctly handles such ``dirty data". After manually correcting for these dataset flaws, we find that only 37 were genuine model failures. Consequently, re-evaluating on a ``clean" subset improves our model's execution accuracy to 96.5\% on the dev set and 97.6\% on the test set (Table~\ref{tab:spider-result}, ``No Gold Errors" row). This finding highlights how benchmark quality issues can mask true model performance, a problem that appears widespread, as a preliminary analysis on a 10\% stratified sample of the BIRD dev set indicates over 30\% of its items contain similar Gold Errors, corroborating recent studies \citep{shen2025study, yang2025automated}. Further details of our Spider test set analysis are available in the supplementary material due to space limitations.

\subsection{Ablation Study}
Table~\ref{bird-ablation} presents our ablation study on the BIRD dev set. ``w/o Planner/SQLGenerator" indicates replacement with MAC-SQL's modules, while other variants remove the specified component. ``w/o Human-like CoT" indicates replacing the chain-of-thought prompting with a standard zero-shot prompt. The results confirm that each component of the GBV-SQL framework is integral, as removing any one of them degrades overall performance. Notably, ablating the SQLChecker, which handles final formatting and executability analysis, causes the most significant performance drop. This underscores the profound impact of query executability and proper formatting on the current execution based (EX) evaluation paradigm.
\begin{table}[t]
\centering
\small 
\setlength{\tabcolsep}{4pt} 
\begin{tabular}{lcccc}
\toprule
\textbf{Model} & \textbf{Simple} & \textbf{Mod.} & \textbf{Chall.} & \textbf{Total} \\
\midrule
GBV-SQL + Deepseek-v3 & 69.51 & 54.62 & 50.69 & \textbf{63.23} \\
\quad w/o Planner & 68.76 & 53.98 & 45.83 & 62.13 ($\downarrow$) \\
\quad w/o SQLGenerator & 67.78 & 53.76 & 48.61 & 61.73 ($\downarrow$) \\
\quad w/o Human-like CoT & 66.70 & 52.90 & 51.39 & 61.08 ($\downarrow$) \\
\quad w/o SQL2TextValidator & 68.54 & 52.69 & 47.92 & 61.80 ($\downarrow$) \\
\quad w/o SQLChecker & 66.49 & 49.89 & 47.22 & 59.65 ($\downarrow$) \\
\bottomrule
\end{tabular}
\caption{Ablation of GBV-SQL on the BIRD's dev set. Mod. and Chall. stand for Moderate and Challenging, respectively.}
\label{bird-ablation}
\end{table}
\subsection{Discussion}
The standard Text2SQL evaluation compares the execution results of predicted queries against gold queries, a process highly susceptible to dataset quality. For instance, an ambiguous natural language question can render a single gold SQL query insufficient for reliable judgment. Likewise, a correct question paired with a flawed gold SQL will still yield an inaccurate EX score. More critically, database-level errors can cause substantial deviations in evaluation. Therefore, we strongly urge the community to place greater emphasis on the integrity of Text2SQL benchmarks, adopting a more rigorous approach to their creation and maintenance. Benchmarks should not be treated as static, infallible standards, but as dynamic entities that require continuous validation, correction, and versioning.

\section{Conclusion}
This paper investigates the problem of semantic inconsistency in Text2SQL, proposing GBV-SQL, a novel multi-agent framework using Guided Generation and SQL2Text Back-translation Validation. The method achieves a 5.8\% accuracy improvement on the BIRD benchmark. To better assess our model's capabilities, we analyze the Spider dataset using our proposed ``Gold Error" typology. This analysis reveals that most prediction failures stem from benchmark flaws rather than model deficiencies. This re-evaluation on valid subsets elevates GBV-SQL's effective accuracy to 96.5\% and 97.6\% on the Spider dev and test sets, respectively. Our work, therefore, offers a promising framework for semantic fidelity and a methodology for benchmark curation, suggesting future research directions in automated dataset validation and the extension of this validation mechanism to other complex reasoning tasks.

\appendix

\bigskip

\bibliography{aaai2026}


\end{document}